\documentclass{article}
\usepackage{ijcai19}

\usepackage{times}

\usepackage{url}
\usepackage[draft]{hyperref}
\usepackage[utf8]{inputenc}
\usepackage[small]{caption}
\usepackage{graphicx}
\usepackage{amsmath,amssymb,amsfonts}
\usepackage{booktabs}
\usepackage[ruled,vlined]{algorithm2e}
\usepackage{subfig}
\usepackage{latexsym} 

\newcommand{\taskprefix}{Q}
\newcommand{\tasksize}{\normalsize}
\newenvironment{taskenv}[1]{\begin{list}{{\tasksize\sc \theenumi.}}{\usecounter{enumi}
      \settowidth{\labelwidth}{{\tasksize\sc \taskprefix#1-99}}
      \setlength{\leftmargin}{\labelwidth}
}}{\end{list}}
\newcounter{task}
\newcommand{\btask}{\begin{taskenv}{\taskprefix}\setcounter{enumi}{\value{task}}\renewcommand{\theenumi}{\taskprefix$_{\arabic{enumi}}$}}
\newcommand{\etask}{\setcounter{task}{\value{enumi}}\renewcommand{\theenumi}{\arabic{enumi}.}\end{taskenv}}

\title{neuralRank: Searching and ranking ANN-based model repositories}

\author{
Nirmit Desai$^1$\and
Linsong Chu$^1$\and
Raghu K. Ganti$^1$\and
Sebastian Stein$^2$\and
Mudhakar Srivatsa$^1$\\
\affiliations
$^1$IBM T. J. Watson Research Center, Yorktown Heights, NY, USA\\
$^2$University of Southampton, Southampton, UK\\
\emails
\{nirmit.desai, lchu, rganti, msrivats\}@us.ibm.com,
ss2@ecs.soton.ac.uk
}

\begin{document}

\maketitle

\begin{abstract}
Widespread applications of deep learning have led to a plethora of
pre-trained neural network models for common tasks. Such models are
often adapted from other models via transfer learning. The models
may have varying training sets, training algorithms, network
architectures, and hyper-parameters. For a given application, what is
the most suitable model in a model repository? This is a critical
question for practical deployments but it has not received much
attention. This paper introduces the novel problem of searching and
ranking models based on suitability relative to a target dataset and
proposes a ranking algorithm called \textit{neuralRank}. The key idea
behind this algorithm is to base model suitability on the
discriminating power of a model, using a novel metric to measure
it. With experimental results on the MNIST, Fashion, and CIFAR10
datasets, we demonstrate that (1) neuralRank is independent of the
domain, the training set, or the network architecture and (2) that the
models ranked highly by neuralRank ranking tend to have higher model
accuracy in practice.
\end{abstract}

\section{Introduction}
Deep neural networks continue to evolve rapidly and their application
domains are also becoming widespread ranging from image / video
classification and natural language processing to agriculture and
healthcare. With the increased availability of tutorials and easy
access to platforms such as TensorFlow and PyTorch for creating neural
networks, there has been a steep rise in the number of architectures
and corresponding models. For example, the official TensorFlow model
repository\footnote{https://tfhub.dev/}, TensorFlow Hub has several
hundred models.  The open source community has tens of thousands of
models that are available for developers to consume. Some examples of
such open source models include
\href{https://github.com/BVLC/caffe/wiki/Model-Zoo}{Caffe Model zoo},
\href{https://modeldepot.io/browse}{Model Depot}, and
\href{https://github.com/Cadene/pretrained-models.pytorch}{PyTorch
  Pre-Trained models}.

A key challenge in training neural networks is the need for labeled
data, which is difficult to obtain~\cite{yokinski:14}, leading to the
rise of {\it transfer learning}, the idea of reusing layers from
pre-trained models. Transfer learning is a key field of deep learning
that shows that general features learned from vast amounts of labeled
datasets available to a few organizations can be reused by those that
lack access to large labeled datasets. However, a key issue that
developers face is how to choose the right pre-trained model. A naive
mechanism is to tag the models using keywords and let the user search
on these keywords. For example, a search for ``image classification''
in TFHub provides 700 model results. A typical developer uses a
combination of keyword search and other factors such as the depth of
the architecture, the number of parameters to retrain, the input size
of the images, and the input dataset on which the pre-trained model
was generated.  Based on these factors, one can choose multiple models
from a domain and evaluate them in a brute-force manner on the target
dataset based on suitable measures of model performance, e.g.,
accuracy. However, such an approach has two major shortcomings: (1)
the set of labels used for training the pre-trained model may be
different from those in the target dataset, rendering meaningless any
evaluation of models based on predicted labels, and (2) models trained
on domains different from the target domain may be more suitable but
finding those via brute-force is not scalable.  Such an haphazard
approach to searching for the right model seriously hinders mainstream
applications of deep learning.

In this paper, we tackle the problem of finding the right model
through a search and rank approach, analogous to the web search
problem. As such, Internet search engines became dominant because of
their capability to search and return a ranked list of relevant
results. We develop a simple yet elegant solution to the problem of
search and rank in the context of identifying the right pre-trained
model for a given deep learning task. As far as the authors are aware,
this is the first paper that addresses the problem of {\it model
  ranking} in neural networks. Specifically, we define the model
ranking problem as follows: Given a set $\mathbb{M} = \{M_1, M_2,
... , M_Z\}$ of models, and a target dataset $T = \{\langle X_1,Y_1
\rangle, \langle X_2, Y_2 \rangle, ... , \langle X_T,Y_T \rangle\}$,
rank the $Z$ models based on a relevancy metric, where the relevancy
metric identifies the ``best'' model for the classification task on
the target labeled dataset.  Given the heterogeneity of training
algorithms, network architectures, and hyper-parameters, we need to
examine approaches that are generally applicable, in spite of the
heterogeneity.

Our approach to model ranking is simple yet powerful as we observe in
the rest of this paper.  We adopt cluster quality as the main
relevancy metric for ranking the models, based on outputs at each
layer of the neural network produced from transformed outputs of the
preceding layers. Hence, when the target dataset is processed through
an existing pre-trained model, outputs at each layer correspond to a
set of clusters, one cluster per label in the target dataset.  We show
that the cohesiveness and separation of the clusters is independent of
the label set, training algorithm, network architecture, or
hyper-parameters used during training. Further, we show that the
well-known cluster quality metric of Silhouette's coefficient
\cite{rousseeuw-silhouette-1987} that is also computationally
efficient ($O(n^2)$ in the number of samples in the target dataset) is
an accurate predictor of model suitability.

Armed with these insights, this paper proposes {\bf neuralRank} -- an
algorithm for ranking neural-network models that is based on a measure
of cluster quality and assumes no knowledge of how the models are
trained or what domain they belong to. In the neuralRank algorithm,
latent-space projections of a target dataset are first computed. Next,
for meaningful distance measurements, the dimensionality of the
projections is reduced via the well-known technique of principal
component analysis (PCA). Lastly, a Silhouette's coefficient on the
PCA projections is computed using the labels in the target
dataset. Higher scores imply greater cluster quality and hence greater
model suitability.

We evaluate neuralRank by building a model zoo from the MNIST,
Fashion, and CIFAR10 datasets with each model having a different
subset of the class labels to emulate domain and label set
differences.  The experiments are designed to investigate three main
questions: (a) how well does the neuralRank ranking match actual model
performance on the target dataset, (b) is neuralRank independent of
domains, network architecture, and label sets, and (c) how sensitive
is neuralRank to the chosen dimensionality of PCA.  The results
indicate that neuralRank ranking accurately predicts how a model may
perform on the target dataset, even when the target dataset is from a
completely different domain. Further, neuralRank is shown to be
independent of network architecture and label sets of pre-trained
models as well as robust across a wide range of PCA dimensionality.
Lastly, an explanation of why neuralRank approach works is provided
via a visualization of the latent-space.

In summary, this paper makes the following contributions: it (1)
introduces the novel and significant problem of model search, (2)
formulates the problem of model search in terms of cluster quality,
and (3) proposes and evaluates the computationally efficient yet
powerful metric of Silhouette's coefficient for model ranking. The
rest of this paper is organized as follows. Section~\ref{sec:problem}
formulates the problem of model search and {\mbox
  Section~\ref{sec:algo}} details the neuralRank
algorithm. Experimental results are presented in
Section~\ref{sec:eval} with Section~\ref{sec:related} describing
related works. Lastly, Section~\ref{sec:conclude} concludes the paper
with ideas for future work.



\section{Model Search Problem}
\label{sec:problem}
\begin{figure*}[htb!]
  \includegraphics[width=\textwidth]{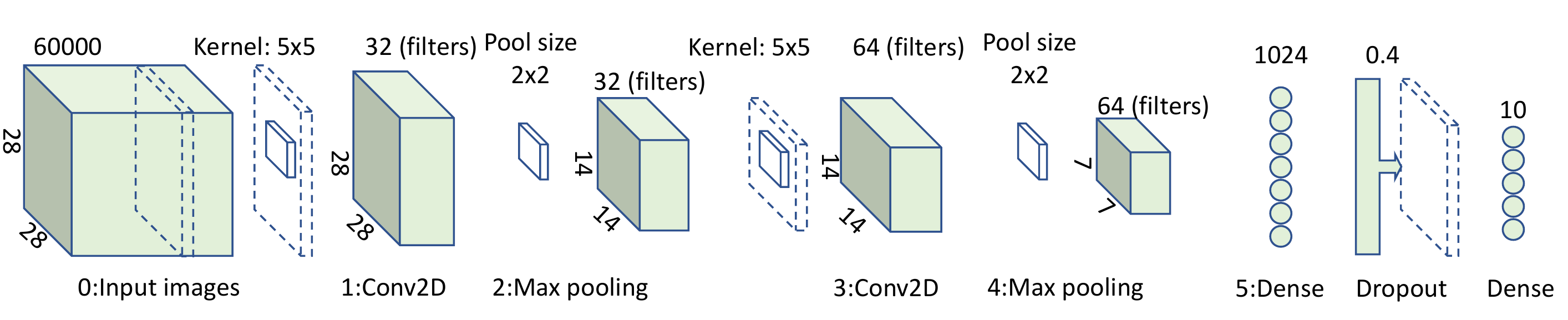}
  \caption{CNN network architecture used for all experiments}
  \label{fig:model}
\end{figure*}
Given a set $\mathbb{M} = \{M_1, M_2, \ldots, M_Z\}$ of models, and a
target dataset $T = \{\langle X_1,Y_1 \rangle, \langle X_2, Y_2
\rangle, \ldots, \langle X_T,Y_T \rangle\}$, rank the $Z$ models based
on a relevancy metric that identifies the ``best'' model for the
classification task on the target labeled dataset.

With infinite resource and a large enough target dataset, an obvious
approach would be applying transfer learning to retrain all model zoo
models or even train new models from scratch. However, given the
scarcity of labeled target datasets as well as the resource costs,
such an approach does not scale. Hence, there is need for a
computationally efficient approach to estimate model performance on a
target dataset without having access to a large number of labeled
samples. Basing model suitability on the discriminating power of
models and a computationally efficient metric to measure the
discriminating power is the key contribution of this paper. We
formally define the problem and the metric in the following.

In general, a neural network consists of layers $0 \le n \le N$, where
$N$ is the number of parameterized layers in the network. Neurons in
layer $n$ produce latent-space output $L^n$ by applying a pre-defined
function $f_n$ to the outputs of the previous layer $L^{n-1}$, thus
$L^n = f_n(L^{n-1})$. In general, the functions implement a variety of
transformations, including but not limited to convolutions, weighted
sum with bias, normalizations, and ReLU activations. $L^0$ is defined
separately as the feature input vector $X_i$. By $f_n(X_i)$, we denote
output $L^n_i$ after the input $X_i$ is processed through all layers
before $n$. Similarly, $L^N$ is the same as the predicted labels
$Y^{'}$ with $Y$ being the ground-truth labels.

With $K$ classes, $T$ labeled inputs in the target dataset, centroid
$C_k^n$ on the outputs of layer~$n$ of class $k$ is defined as follows.

\begin{equation} \label{eq:centroids}
C_k^n = \sum\limits_{i=1}^{T} \frac{L^n_i | Y_i = k}{T}
\end{equation}

Intuitively, a discriminating model should place similar inputs closer
together and as far away from other clusters as possible.  We apply
Silhouette's coefficient $SC$ on the layer $n$ outputs $L^n$
defined as follows in terms of the intra-cluster (cohesion) and
inter-cluster distance (separation) measures $a$ and $b$,
respectively.

\begin{equation} \label{eq:sc}
SC(L^n, Y) = \sum\limits_{i=1}^{T} \frac{\frac{b(L^n_i) - a(L^n_i)}{max(a(L^n_i),b(L^n_i))}}{T}
\end{equation}

Mean intra-cluster distance $a$ is defined relative to a point $L^n_i$
and all other points $L^n_j$ in the same class as $L^n_i$ as shown in
Equation~\ref{eq:a}. Mean intra-cluster distance $b$ for a given point
$L^n_i$ is simply the distance to the centroid of the nearest class as
shown in Equation~\ref{eq:b}. $C_k^{n*}$ indicates the centroid of the
closest class to $L^n_i$. 

\begin{equation} \label{eq:a}
a(L^n_i) = \sum\limits_{j=1}^{T} \frac{dist(L^n_i, L^n_j)| Y_i, Y_j = k, i \neq j}{T}
\end{equation}

\begin{equation} \label{eq:b}
b(L^n_i) = dist(L^n_i - C_k^{n*})| Y_i \neq k
\end{equation}

As $a$ and $b$ need to be computed for all points $L^n_i \in L^n$,
computing SC involves $T^2$ distance computations on the latent-space
projections $L^n$. A variety of distance measures $dist$ can be
applied, we use cosine distance in this paper due to its robustness in
high-dimensional spaces.  An ideal model generates latent-space
projections $L^n$ that maximize $b$ and minimize $a$, yielding SC
score close to $1$.  When a model cannot discriminate well across
classes in $X$, $b$ and $a$ are similar, yielding SC score close to
$0$.  In the worst case, a model may discriminate within a class but
not across classes, yielding a negative SC score.


\section{neuralRank Algorithm}
\label{sec:algo}
\begin{algorithm}[htb!]
\caption{neuralRank algorithm}
\label{algo:rank}
\DontPrintSemicolon 
\SetAlgoLined
\KwIn{Target dataset $(X, Y)$, Models $\mathbb{M}$}
\KwIn{Layer $n$, Projection dimensionality $D$}
\KwOut{Sorted SC scores $SC_M$ for all $M \in \mathbb{M}$}
 $SC_{list} \gets \varnothing$\;
 \ForEach{$M \in \mathbb{M}$}{
   $N \gets M$.layers\;
   \If{$n \leq N$}{
     $L^{n} \gets M_{n}(X)$\;
     $PCA_{n} \gets PCA(L^{n}, D)$\;
     $SC_{M} \gets SC(PCA_{n}, Y)$\;
     $SC_{list} \gets SC_{list} \cup SC_{M}$\;
   }
   \Else{
     Raise Error
   }
 }
 
 \Return{Sort($SC_{list}$)}\;
\end{algorithm}
The algorithm involves computing SC score on latent-space projections
at a chosen layer $n$ for all models in the model zoo, given a target
dataset $(X, Y)$. Later, we show via experiments that typically,
choosing the last dense layer works well as it captures the richest
features in the inputs.  No knowledge of how the models were trained or
network architecture of models is assumed. 

Algorithm~\ref{algo:rank} outlines the main steps of the neuralRank
algorithm. Apart from the target dataset and the model repository, two
parameters are needed: (1) $n$ is the chosen latent-space layer and
(2) $D$ is the dimensionality to which $L^n$ needs to be reduced
to. Latent-space outputs commonly have a large number of dimensions
and distance measures do not work well on high-dimensional vectors. To
overcome this, PCA is applied on the outputs $L^n$ to reduce the
dimensionality down to $D$ and SC is measured on the reduced PCA
projections. 


\section{Experimental Evaluation}
\label{sec:eval}
We evaluate neuralRank on the datasets from three separate domains:
MNIST hand-written digit images~\cite{lecun-mnist-2010}, Fashion
apparel images~\cite{xiao-fashion-mnist-2017}, and CIFAR10 birds,
animals, and vehicles images~\cite{cifar10}. Both MNIST and fashion
have $28 \times 28$ grey scale images whereas CIFAR10 has $32 \times
32$ color images that we resize to $28 \times 28$ grey scale.  Each of
the datasets have 10 classes with 60K images in the training set and
10K images in the validation set. The validation sets from each of the
datasets are used in all of the experiments. In the first set of
experiments, the same CNN-based network architecture as shown in
Figure~\ref{fig:model} is used for training the models for each
dataset. Later, we also experiment with a different architecture. A
dropout rate of $0.4$ is applied on the last dense layer for
regularization. All models are trained with learning rate of $0.001$
and cross-entropy loss. By default, $D=10$ is used in the PCA
reduction step.
\begin{table}
\centering
\begin{tabular}{lll}
\hline
Model name  & Training Dataset & Training Classes \\
\hline
$M_{10}$       & MNIST  & all     \\
$M_{0-4}$       & MNIST  & 0-4      \\
$M_{5-9}$    & MNIST  & 5-9     \\
$F_{10}$       & Fashion  & all     \\
$F_{0-4}$       & Fashion  & 0-4      \\
$F_{5-9}$    & Fashion  & 5-9     \\
$C_{10}$       & CIFAR10  & all     \\
$C_{0-4}$       & CIFAR10  & 0-4      \\
$C_{5-9}$    & CIFAR10  & 5-9     \\
\hline
\end{tabular}
\caption{Model zoo: 9 models, 3 datasets}
\label{tab:zoo}
\end{table}

To emulate a model zoo with models having varying degree of domain
differences, three training subsets are created with samples from: (1)
all classes, (2) first five classes, and (3) the remaining
classes. Table~\ref{tab:zoo} describes all the models in the model
zoo. In the following, we investigate these questions via experiments.
\btask
\item\label{q:1} What latent-space layer should the neuralRank scores
  be based on?
\item\label{q:2} Is neuralRank independent of domains?
\item\label{q:3} How well does the neuralRank ranking match actual model performance?
\item\label{q:4} How sensitive are the results to the choice of $D$ in the PCA dimensionality reduction step?
\item\label{q:5} Is neuralRank independent of network architecture?
\item\label{q:6} How sensitive are the results to the dimensionality of the chosen layer?
\etask
%
\subsection{\ref{q:1}: Latent-space layers}
\label{sec:layers}
%
\begin{table}[htb!]
\centering
\begin{tabular}{lcccccc}
\hline
          & $L^0$ & $L^1$ & $L^2$ & $L^3$ & $L^4$ & $L^5$  \\
\hline
$M_{10}$  & 0.346 & \textbf{0.361} & \textbf{0.396} & \textbf{0.480} & \textbf{0.548} & \textbf{0.784}\\
$M_{0-4}$ & 0.346 & \textbf{0.369} & \textbf{0.409} & \textbf{0.477} & \textbf{0.544} & \textbf{0.821}\\
$M_{5-9}$ & 0.346 & \textbf{0.360} & \textbf{0.393} & \textbf{0.391} & \textbf{0.455} & \textbf{0.463}\\
$F_{10}$  & 0.346 & 0.306 & 0.343 & 0.275 & 0.359 & 0.248\\
$F_{0-4}$ & 0.346 & 0.328 & 0.366 & 0.257 & 0.356 & 0.276\\
$F_{5-9}$ & 0.346 & 0.276 & 0.315 & 0.272 & 0.350 & 0.210\\
$C_{10}$  & 0.346 & 0.329 & 0.367 & 0.313 & 0.368 & 0.293\\
$C_{0-4}$ & 0.346 & 0.351 & 0.380 & 0.337 & 0.394 & 0.315\\
$C_{5-9}$ & 0.346 & 0.324 & 0.361 & 0.325 & 0.389 & 0.303\\
\hline
\end{tabular}
\caption{SC scores based on projections at $L^0$--$L^5$}
\label{tab:sc1}
\end{table}
We test neuralRank with latent-space projections at each layer to find
the layer at which the scores of the most suitable models exhibit the
sharpest contrast with the rest of the models. Silhouette's
coefficient (SC) defined in Eq~\ref{eq:sc}, is applied to rank all
models in the zoo, assuming a target dataset of MNIST inputs with
classes $0-4$.  Table~\ref{tab:sc1} shows SC scores across $L^0-L^5$
(layer indices from Figure~\ref{fig:model}) with top-3 ranked models
highlighted. The MNIST models consistently achieve the highest scores,
which is not surprising given the target dataset. Importantly, the SC
scores on first few layers do not exhibit much contrast whereas the
last dense layer shows the sharpest contrast. This is expected because
the level of abstraction increases with higher layers. We shall focus
on $L^5$ for subsequent experiments.
\subsection{\ref{q:2}: Domain independence}
\label{sec:domain}
%
\begin{table*}[htb!]
\centering
\begin{tabular}{ccccccccc}
\hline
$M_{10}$ & $M_{0-4}$ & $M_{5-9}$ & $F_{10}$ & $F_{0-4}$ & $F_{5-9}$ & $C_{10}$ & $C_{0-4}$ & $C_{5-9}$\\
\hline
0.099   & 0.081    & 0.061    & \textbf{0.499}   & \textbf{0.540}    & 0.082    & \textbf{0.212}   & 0.182    & 0.180\\
\hline
\end{tabular}
\caption{SC scores on $L^5$ projections of Fashion inputs with classes
  $0-4$}
\label{tab:sc2}
\end{table*}
\begin{table*}[htb!]
\centering
\begin{tabular}{ccccccccc}
\hline
$M_{10}$ & $M_{0-4}$ & $M_{5-9}$ & $F_{10}$ & $F_{0-4}$ & $F_{5-9}$ & $C_{10}$ & $C_{0-4}$ & $C_{5-9}$\\
\hline
0.862   & 0.871    & 0.859    & \textbf{0.942}   & \textbf{0.943}    & 0.875    & \textbf{0.892}   & 0.886    & 0.883\\
\hline
\end{tabular}
\caption{Accuracy of transfer learned models on the target dataset Fashion $0-4$}
\label{tab:sc3}
\end{table*}
If neuralRank is indeed domain independent, rankings should not
necessarily follow domain boundaries. Hence, we should be able to find
cases where models outside of the domain of the target dataset are
found to be more suitable than the ones matching the domain. Results in
Table~\ref{tab:sc1} follow along the domain boundaries and hence do
not help answer this question.  In the next experiment, we choose a target
dataset of Fashion inputs with classes $0-4$ and observe the SC scores
on layer $5$. Table~\ref{tab:sc2} shows the results with top-3 ranked
models highlighted.  The top-2 ranks go to $F_{0-4}$ and $F_{10}$ and
follow along the domain boundary. However, the third rank goes to
$C_{0-4}$, which is a completely different domain (animals and
vehicles) and has a clearly better score than $F_{5-9}$. One
explanation is that CIFAR10 being a richer and noisier dataset, CIFAR
models have been forced to learn to discriminate with richer features
and such discriminatory power is domain independent.

\subsection{\ref{q:3}: Ranking and model performance}
\label{sec:performance}
%
\begin{figure}[htb!]
  \includegraphics[width=\columnwidth]{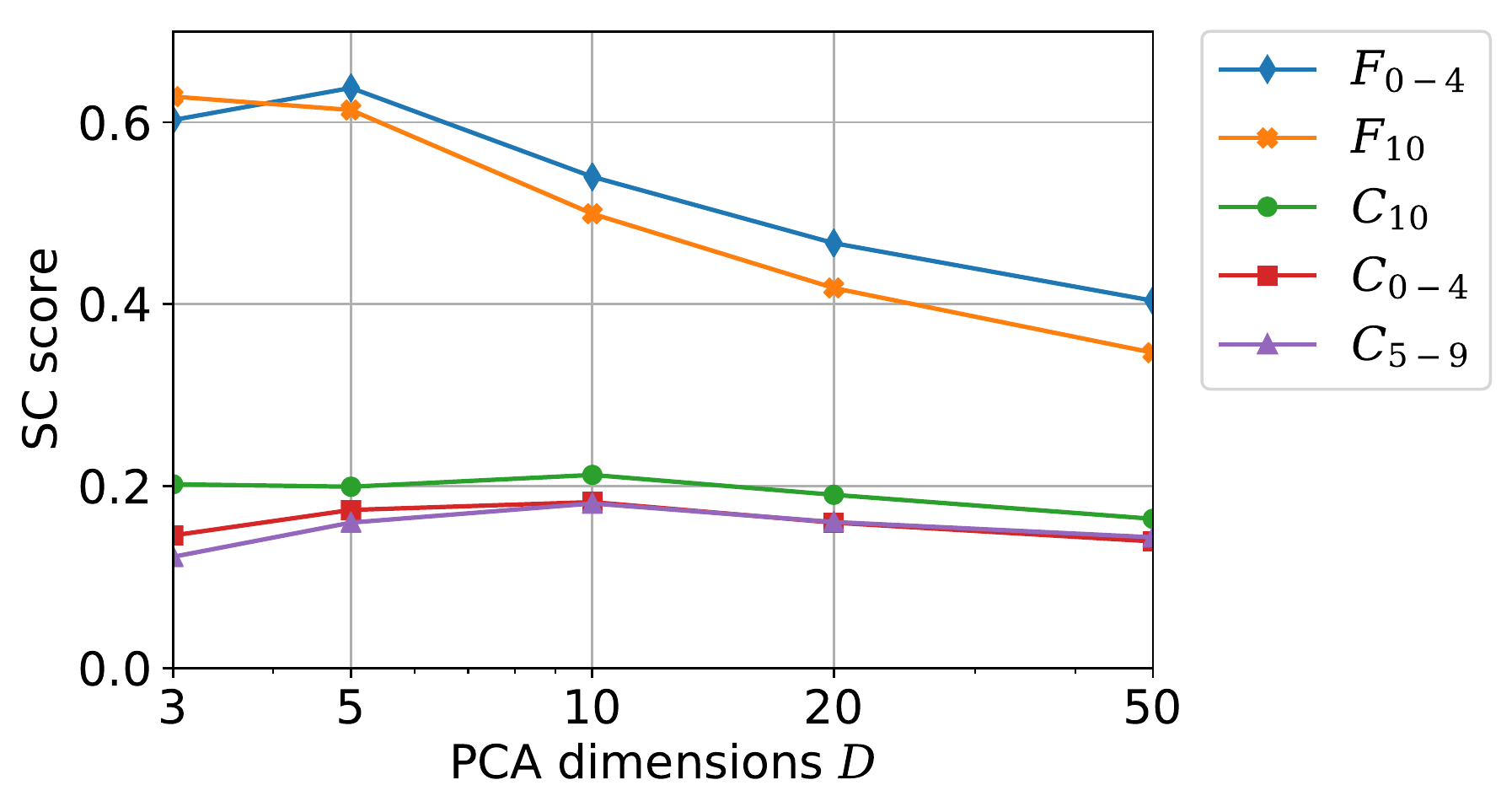}
  \caption{SC scores of top-5 models from Table~\ref{tab:sc2} on $L^5$ with varying PCA dimensions $D$}
  \label{fig:pca}
\end{figure}
Although the above SC score ranking results make sense, whether or not
the high-ranked models actually perform well on the target datasets
remains to be validated. To validate this, with the same target
dataset as in Section~\ref{sec:domain}, i.e., Fashion inputs with
classes $0-4$, we apply the standard transfer learning technique of
freezing weights on all layers except the logits layer.  As a result,
logits layer of each of the models in the zoo is retrained to
correspond to the classes in the target dataset. Table~\ref{tab:sc3}
reports the prediction accuracy of each of the transferred models on
the target dataset. The ranking of models based on SC for layer~$5$ in
Table~\ref{tab:sc2} matches well the ranking of models based on
accuracy. This result validates that SC scores can reliably predict
model performance.

\begin{figure*}[htb!]
\begin{tabular}{ccc}
  \includegraphics[width=0.3\textwidth]{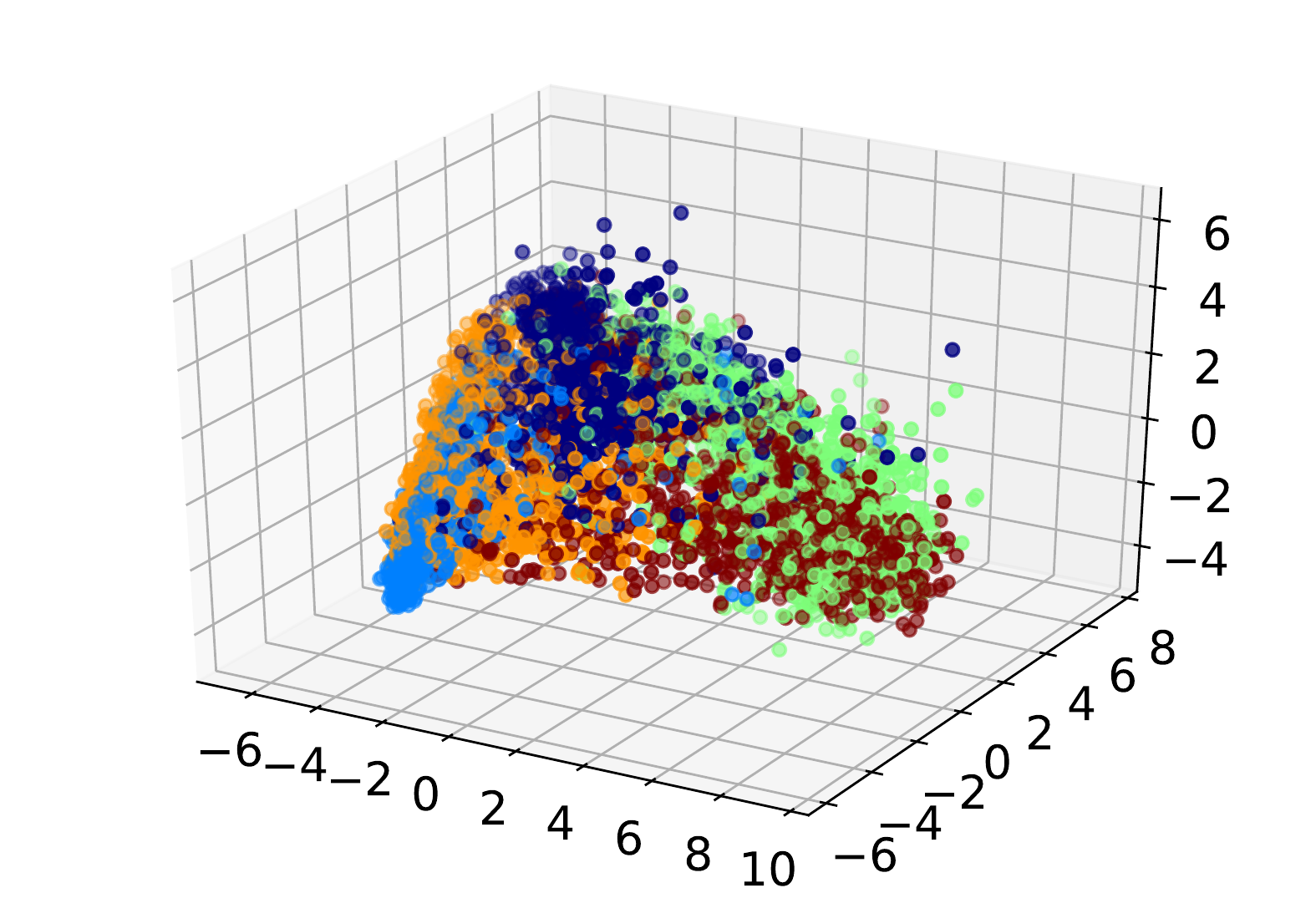}   &   
  \includegraphics[width=0.3\textwidth]{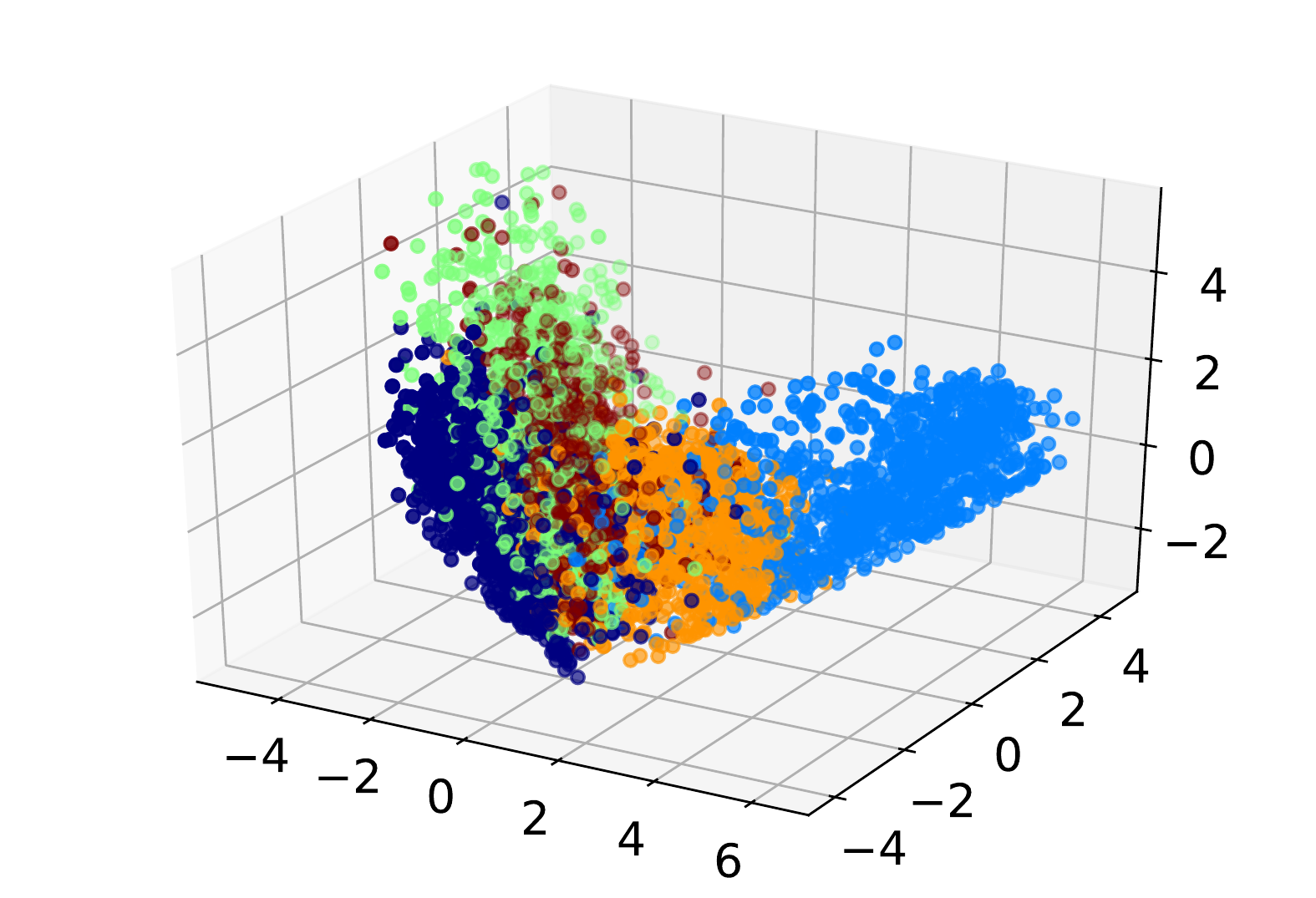}   &
  \includegraphics[width=0.3\textwidth]{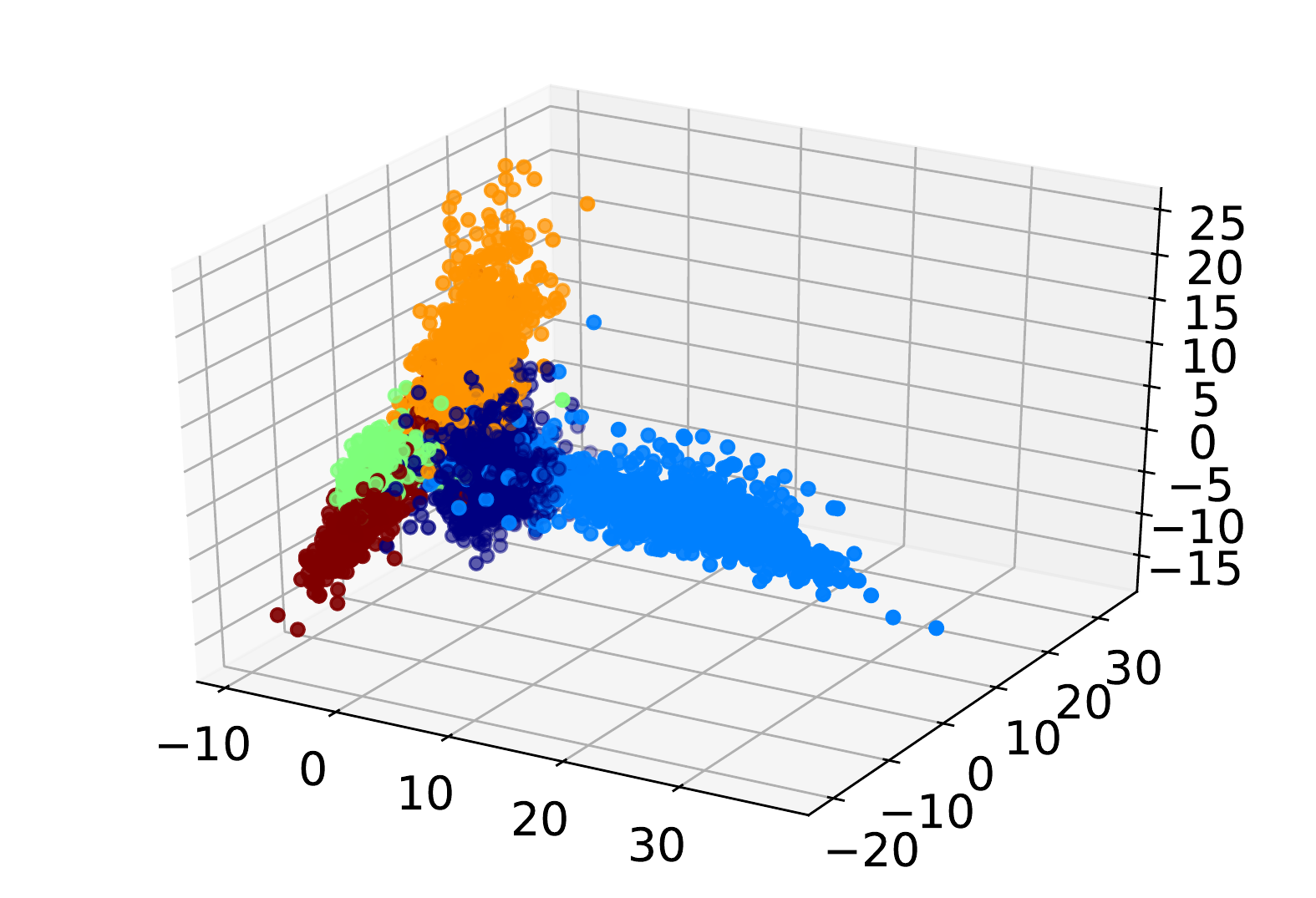}   \\
(a) $F_{0-4}$, $L^0$ & (b) $F_{0-4}$, $L^2$ & (c) $F_{0-4}$, $L^5$ \\[6pt]
  \includegraphics[width=0.3\textwidth]{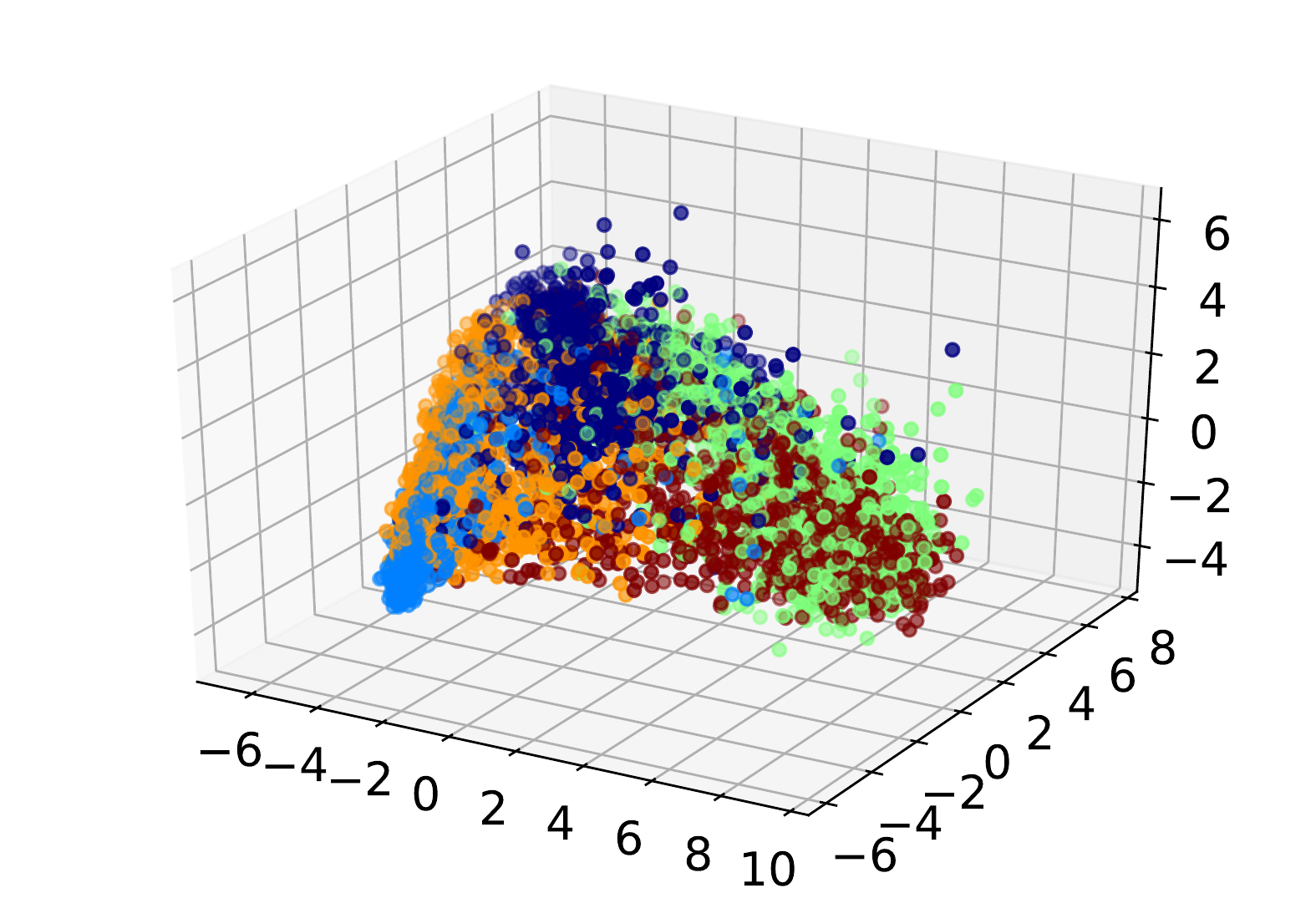}   &
  \includegraphics[width=0.3\textwidth]{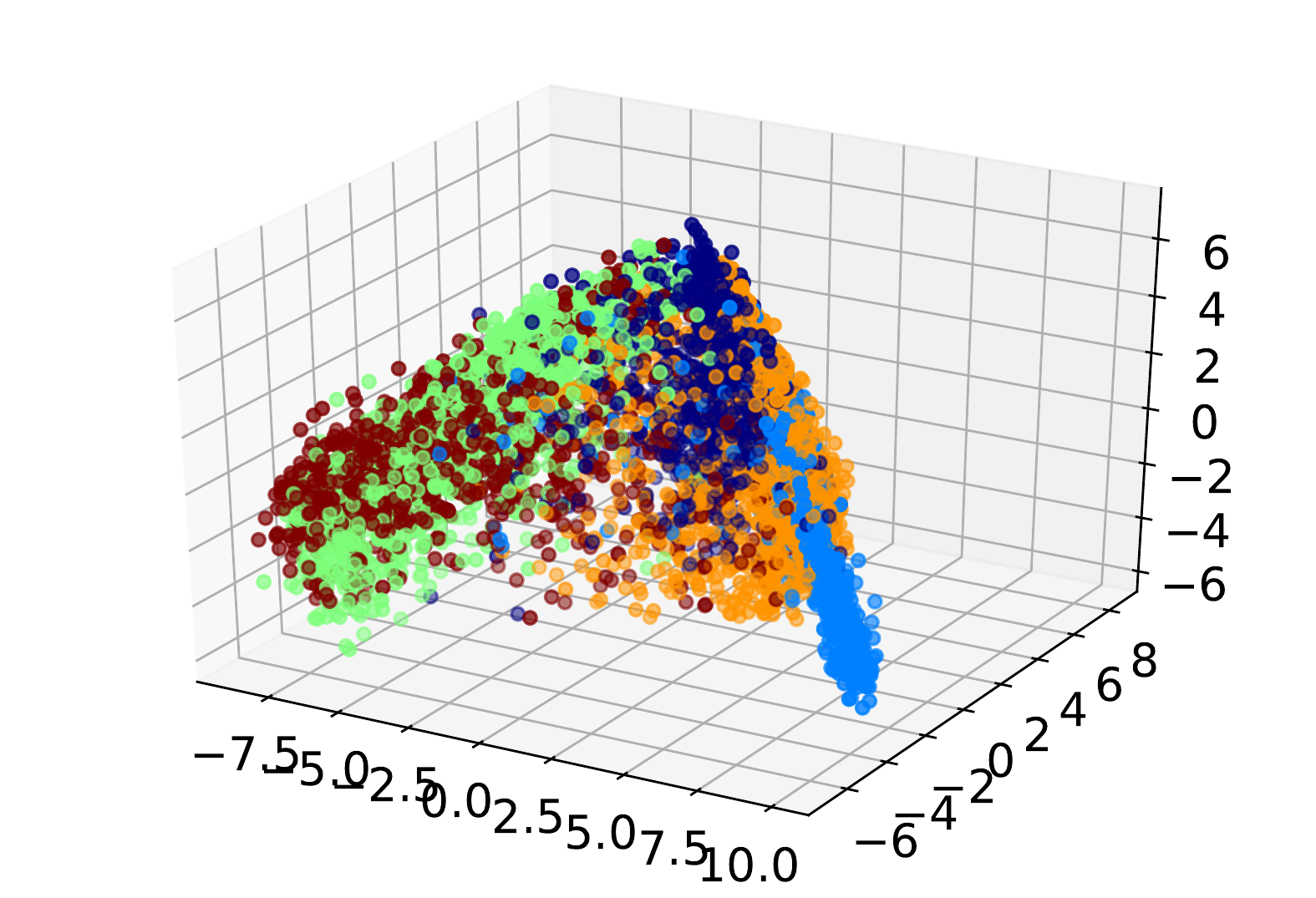}   &
  \includegraphics[width=0.3\textwidth]{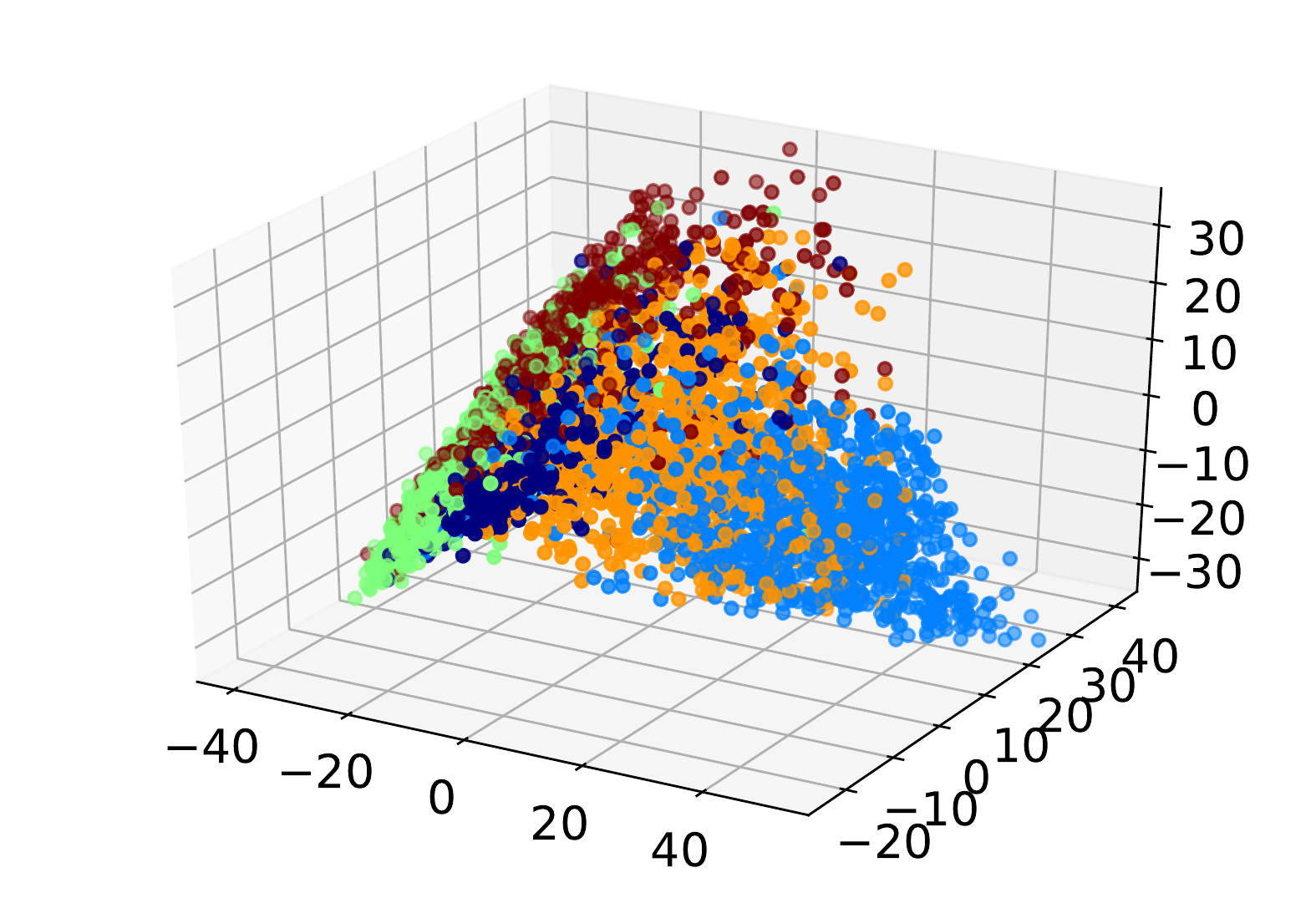}   \\
(d) $C_{10}$, $L^0$ & (e) $C_{10}$, $L^2$ & (f) $C_{10}$, $L^5$ \\[6pt]
  \includegraphics[width=0.3\textwidth]{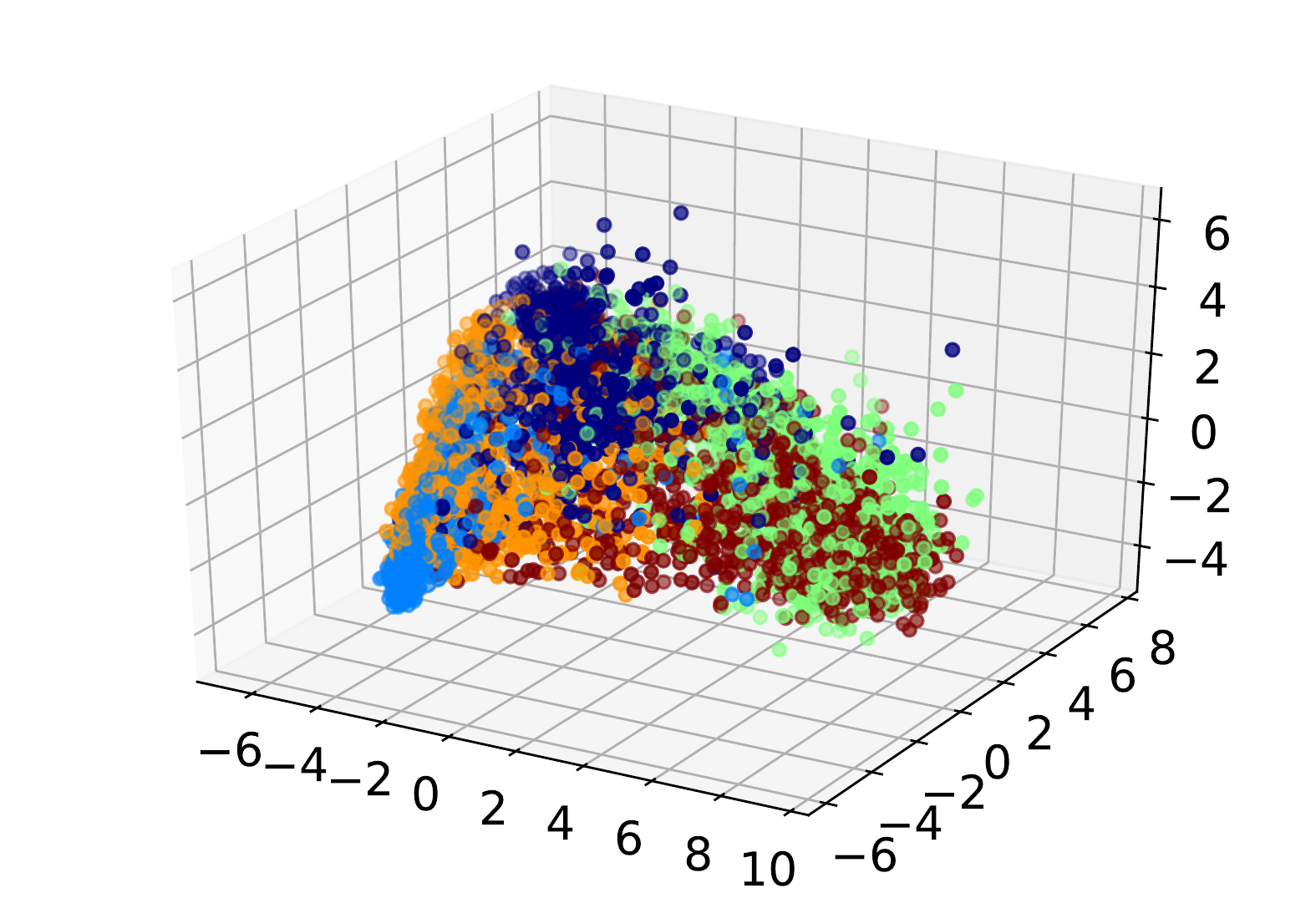}   &
  \includegraphics[width=0.3\textwidth]{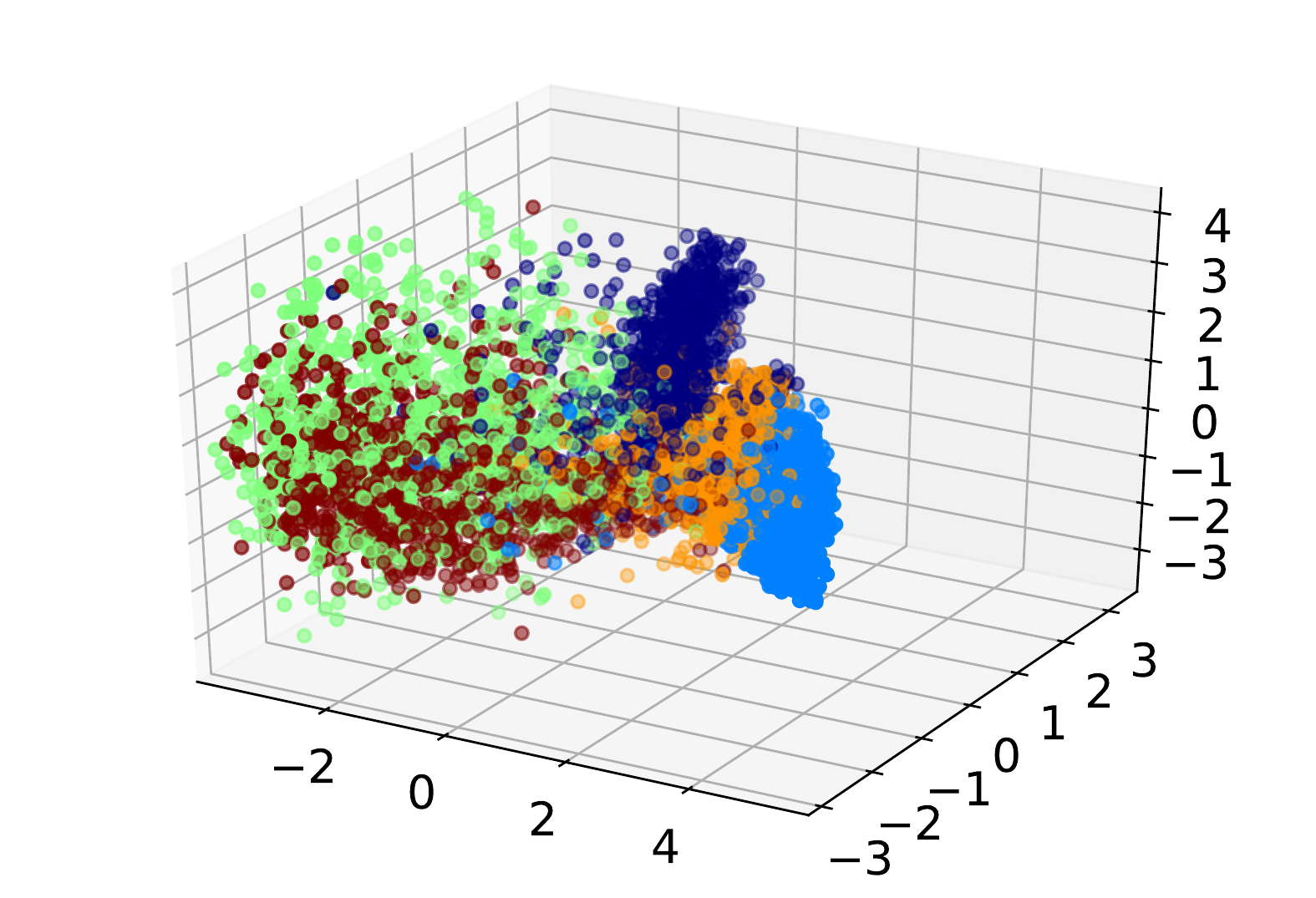}   &
  \includegraphics[width=0.3\textwidth]{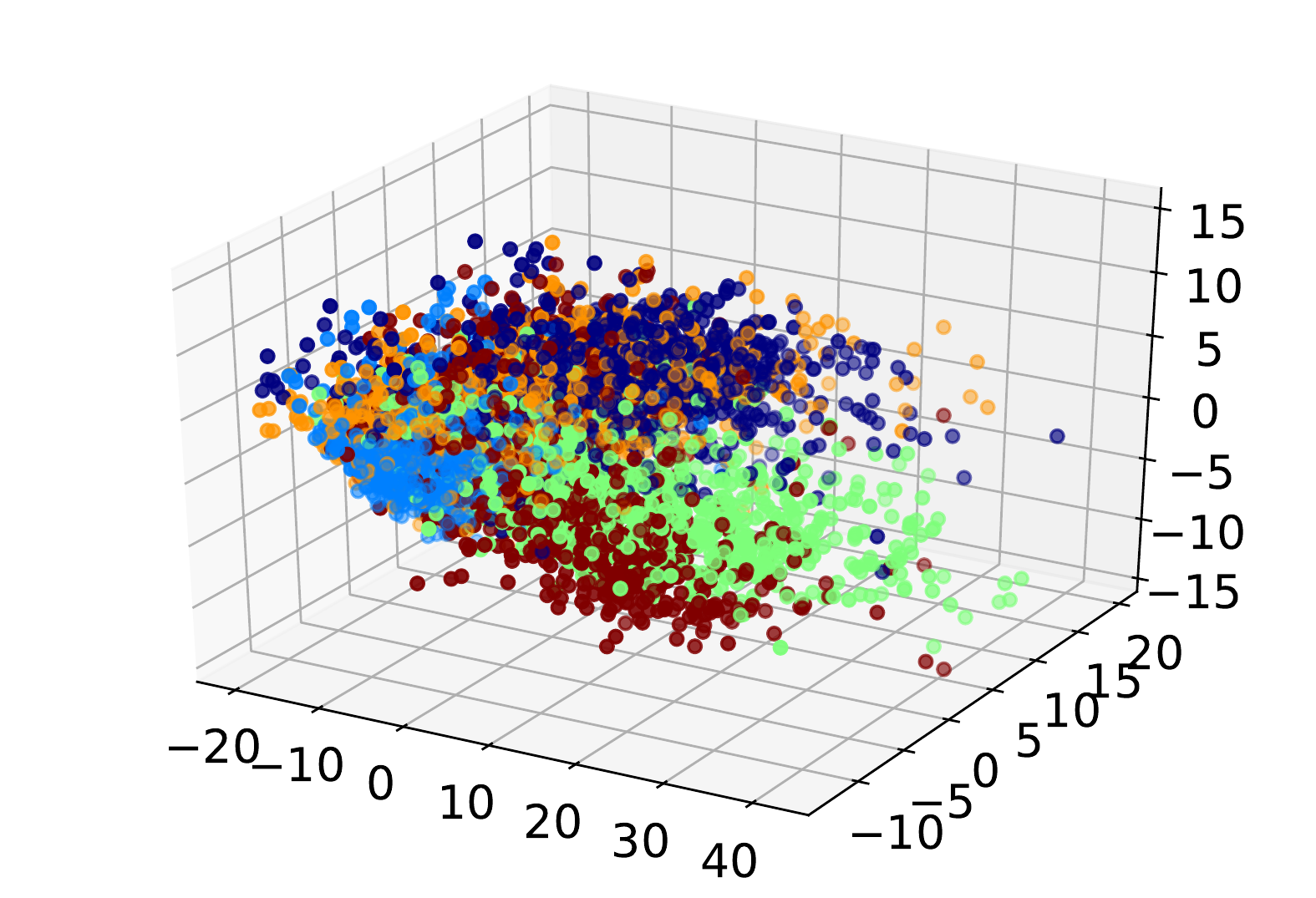}   \\
(g) $F_{5-9}$, $L^0$ & (h) $F_{5-9}$, $L^2$ & (i) $F_{5-9}$, $L^5$ \\[6pt]

\end{tabular}
\caption{3D PCA visualization of latent-space projections of Fashion 0-4 inputs on $F_{0-4}$, $C_{10}$, and $F_{5-9}$}
\label{fig:viz}
\end{figure*}

Although the above results establish the power of SC score
experimentally, explaining why it works is important as well. Since
the SC score measures the discriminatory power of models, higher rank
should correspond to better discrimination. To explain this visually,
Figure~\ref{fig:viz} depicts 3-dimensional PCA of the latent-space
projections $L^0$, $L^2$, and $L^5$ of Fashion inputs with classes 0-4
using $F_{0-4}$, $C_{10}$, and $F_{5-9}$ models (ranks 1, 3, and 7 in
Table~\ref{tab:sc2}). Color indicates class labels.

As expected, for all three models, $L^5$ shows the best discrimination
among classes. More importantly, $F_{0-4}$ (rank 1) shows a much
better discrimination than the other two models. Similarly, $C_{10}$
(rank 3) shows better discrimination than $F_{5-9}$ (rank 7),
explaining why it should rank higher even though it was trained on a
completely different domain.

\begin{table*}[htb!]
\centering
\begin{tabular}{l|ccccccccc}
\hline
& $M_{10}^{l}$ & $M_{0-4}^{l}$ & $M_{5-9}^{l}$ & $F_{10}^{l}$ & $F_{0-4}^{l}$ & $F_{5-9}^{l}$ & $C_{10}^{l}$ & $C_{0-4}^{l}$ & $C_{5-9}^{l}$\\
\hline
MNIST SC score & \textbf{0.671}   & \textbf{0.723}    &  \textbf{0.419}    & 0.226   & 0.231    & 0.189    & 0.301   & 0.305    & 0.304\\
MNIST Accuracy & \textbf{0.998}  & \textbf{0.998}     &  \textbf{0.992}    & 0.987   & 0.975    & 0.977    & 0.986   & 0.982    & 0.986\\
Fashion SC score & 0.174   & 0.134    & 0.107    & \textbf{0.418}   & \textbf{0.436}    & 0.170    & \textbf{0.235}   & 0.195    & 0.212\\
Fashion Accuracy & 0.881 & 0.864 & 0.868 & \textbf{0.939} & \textbf{0.939} & 0.878 & \textbf{0.883} & 0.855 & 0.876\\
\hline
\end{tabular}
\caption{SC scores at $L^3$  and accuracy after transfer learning on lite models}
\label{tab:sc4}
\end{table*}
\begin{table*}[htb!]
\centering
\begin{tabular}{l|cccccccccc}
\hline
                 & $F_{10}$ & $F_{0-4}$ & $C_{10}$ & $C_{0-4}$ & $C_{5-9}$ & $F_{10}^{512}$ & $F_{0-4}^{512}$ & $C_{10}^{512}$ & $C_{0-4}^{512}$ & $C_{5-9}^{512}$ \\
\hline
SC score & \textbf{0.499}  &  \textbf{0.540}  &  \textbf{0.212}  & 0.182  & 0.180 &  \textbf{0.476}  & \textbf{0.561}  & 0.191  &  0.185  & 0.160\\
Accuracy & \textbf{0.940}  &  \textbf{0.940}  &  \textbf{0.893}  & 0.881  & 0.879 &  \textbf{0.946}  & \textbf{0.946}  & 0.851  &  0.863  & 0.868 \\
\hline
\end{tabular}
\caption{SC scores at $L^5$ on Fashion inputs with 0--4 classes and accuracy after transfer learning on models with $512$ neurons in $L^5$}
\label{tab:512}
\end{table*}
%
\subsection{\ref{q:4}: Sensitivity to PCA dimensionality}
\label{sec:pca}
As described in Section~\ref{sec:algo}, a key step before computing SC
scores is to reduce the dimensionality of latent-space projections to
$D$ and avoid distance measurements on high-dimensional
vectors. Hence, it is important to show that the results are not
highly-sensitive to choice of $D$ and choosing a reasonable $D$ is
trivial. Figure~\ref{fig:pca} shows changes to SC scores on $L^5$ with
Fashion inputs having classes 0-4, with varying values of $D$ for the
top-5 models from Table~\ref{tab:sc2}. Line cross-overs correspond to
changes in ranking results. As is evident from the plot, ranking
results do not change across a large range of $D$ values, generally in
the ballpark of $5-50$, which is about two-orders of magnitude fewer
dimensions than the original vector with 1024 dimensions for
$L^5$. The lines do cross-over for very small and very large values of
$D$, which is expected.
%
\subsection{\ref{q:5}: Network architecture independence}
\label{sec:network}
All of the results so far are based on the architecture shown in
Figure~\ref{fig:model} and do not demonstrate network architecture
independence on their own.  In this experiment, we modify the
architecture by dropping the second convolutional and max-pool layers
and train nine more models with the same training set and algorithms
as before. We call this the \textit{lite} models and denote them as
before with the lite annotation, e.g., $M_{10}^{l}$ is a lite version
of $M_{10}$. Next, we repeat the ranking experiments on the new models
with the same target datasets: (1) MNIST inputs with classes $0-4$ and
(2) Fashion inputs with classes $0-4$. Table~\ref{tab:sc4} shows the
SC score on the last dense layer $L^3$ for both target datasets along
with the actual accuracy after re-training the logits layer.  Clearly,
ranking and accuracy results on lite models match previous results,
confirming that regardless of the network architecture, SC score is a
relaible predictor of model performance.  Further, taking the results
of Tables~\ref{tab:sc2}, \ref{tab:sc3}, and \ref{tab:sc4} together,
ranking of all 18 models based on SC score matches the ranking based
on the accuracy results.

\subsection{\ref{q:6}: Dimensionality of the chosen layer}
\label{sec:512}
%
In the above experiments, the last dense layer outputs ($L^5$ in the
original models and $L^3$ in the lite models) had the same
dimensionality of 1024 neurons (as shown in
Figure~\ref{fig:model}). As the dimensionality of the chosen layer may
vary greatly in practice, it is important to study how sensitive
neuralRank is to differences in the dimensioanlity of the chosen
layer.  To test this, we create a model zoo consisting of the top-5
best performing models from Table~\ref{tab:sc2}, modify layer~5 to
have 512 neurons, and train five more models with the modified
layer~5.  Table~\ref{tab:512} shows the SC scores and accuracy results
after transfer learning on all ten models.  The new models with 512
neurons are annotated with $512$, i.e., $F_{10}^{512}$ is the modified
version of $F_{10}$.  As the results show, regardless of the
dimensions in $L^5$, top-5 models according to the SC score are also
the top-5 models based on actual accuracy performance.


\section{Related Work}
\label{sec:related}
This paper is the first to introduce the notion of searching across
models generated by neural networks. As such, we believe that the
closest relevant work is that of transfer learning, where the goal is
to understand which layers are applicable to a new dataset and
``transfer'' these to a ``new'' network. Several papers have explored
this area, a few key ones are~\cite{yokinski:14,cortes:15}. The key
premise of transfer learning is to determine which portions of an
existing neural network can be reused and to what extent. The
reusability depends on the application domain.  For
example,~\cite{yokinski:14,sun:14} show that many of the initial
layers of a deep neural network trained for specific image and video
classification tasks are applicable to other tasks (different from the
one that they were trained on). Another recent example is in the
domain of Natural Language Processing, where models such as
BERT~\cite{devlin:18}, ELMo~\cite{peters:2018}, and
ULMFiT~\cite{howard:18} are fine tuned to be used for various text
problems. The success of transfer learning in these domains shows that
certain portions of neural networks can be ``frozen'' and
reused. However, these fall short of addressing which models are the
best suited for the problem at hand. This paper builds on concepts
from transfer learning and shows that we can use a simple cluster
quality metric to determine which components of the neural network
are applicable in a structured manner.

The next relevant work is that of interpretability, where the goal is
to better understand what each layer of the neural network is
``doing''. One of the approaches to interpretability is to use
visualization
~\cite{olah2017feature,olah2018the,Krause:2016,netdissect2017}, where
each of the layers of the neural network is visualized as an
image. This approach is quite useful when working with image data,
which provides some insights into what each layer is capturing. For
example, as is well-illustrated in~\cite{olah2017feature}, the initial
layers are capturing edges, followed by textures and patterns, then
parts and finally objects (leading to classes). One notices that as we
gradually go up the layers, the complexity of the recognition
grows. These visualizations are quite useful for humans to understand
why a network is working as observed and what kind of features it is
learning in each layer. However, the work on visualization does not
quantify the applicability of layers for a different task.


\section{Conclusions}
\label{sec:conclude}
As applications of deep learning models grow, large repositories of
models are springing up.  This paper introduced the problem of
searching and ranking such model repositories. A novel technique for
ranking models is proposed based on Silhouette's coefficient on
latent-space projections. A key new insight is that the discriminatory
power of a model is a reliable indicator of model suitability
regardless of the domain or the training method. The proposed
technique also lends itself well to a visual explanation of rankings.
Through rigorous experiments, the ranking algorithm is shown to work
well across domains, network architectures, and dimensionality of the
chosen latent-space projections. Most importantly, the ranking results
correspond well to the actual model performance after transfer
learning.

Although these are promising results, this paper represents a first
step in solving the broader problem of model search.  A key assumption
needs to be relaxed for solving the broader problem: an unsupervised
technique for model search wherein the target dataset is not
labeled. This is a significant problem because in many applications,
acquiring sufficient amount of labeled data is costly or even
infeasible.  Also, this is a challenging problem because model
suitability is relative to the needs of an application and labels
represent the application requirements, e.g., classifying vehicles in
images versus classifying landmarks.


\bibliographystyle{named}
\bibliography{ijcai19}

\end{document}